\documentclass[sigconf]{acmart}
\setcopyright{none}
\usepackage{balance}
\usepackage[colorinlistoftodos,prependcaption,textsize=tiny]{todonotes}
\usepackage{caption}
\usepackage{tabularx, environ}
\usepackage{titlesec}
\usepackage{float}

\copyrightyear{2020}
\acmYear{2020}
\setcopyright{acmcopyright}
\acmConference[CHIIR '20]{2020 Conference on Human Information Interaction and Retrieval}{March 14--18, 2020}{Vancouver, BC, Canada}
\acmBooktitle{2020 Conference on Human Information Interaction and Retrieval (CHIIR '20), March 14--18, 2020, Vancouver, BC, Canada}
\acmPrice{15.00}

\begin{document}
\fancyhead{}

\author{Ali Ahmadvand}
\affiliation{\institution{Computer Science Department, \\Emory University, Atlanta, GA}}
\email{ali.ahmadvand@emory.edu}

\author{Harshita Sahijwani}
\affiliation{\institution{Computer Science Department, \\ Emory University, Atlanta, GA}}
\email{hsahijw@emory.edu}

\author{Eugene Agichtein}
\affiliation{\institution{Computer Science Department, \\ Emory University, Atlanta, GA}}
\email{eugene.agichtein@emory.edu}

\title{Would you Like to Talk about Sports Now? Towards Contextual Topic Suggestion for Open-Domain Conversational Agents}

\begin{abstract}
 To hold a true conversation, an intelligent agent should be able to occasionally take initiative and recommend the next natural conversation topic. This is a challenging task. A topic suggested by the agent should be relevant to the person, appropriate for the conversation context, and the agent should have something interesting to say about it. Thus, a scripted, or one-size-fits-all, popularity-based topic suggestion is doomed to fail. Instead, we explore different methods for a personalized, contextual topic suggestion for open-domain conversations. We formalize the Conversational Topic Suggestion problem (CTS) to more clearly identify the assumptions and requirements. We also explore three possible approaches to solve this problem: (1) model-based sequential topic suggestion to capture the conversation context (CTS-Seq), (2) Collaborative Filtering-based suggestion to capture previous successful conversations from similar users (CTS-CF), and (3) a hybrid approach combining both conversation context and collaborative filtering. To evaluate the effectiveness of these methods, we use real conversations collected as part of the Amazon Alexa Prize 2018 Conversational AI challenge. The results are promising: the CTS-Seq model suggests topics with 23\% higher accuracy than the baseline, and incorporating collaborative filtering signals into a hybrid CTS-Seq-CF model further improves recommendation accuracy by 12\%. Together, our proposed models, experiments, and analysis significantly advance the study of open-domain conversational agents, and suggest promising directions for future improvements. 
\end{abstract}

\maketitle

\vspace{-0.3cm}
\section{INTRODUCTION}




While the art of conversation can be considered a uniquely human trait~\citep{tomasello2010origins}, ``artificial'' conversational intelligence (Conversational AI) agents have been gaining traction, especially with the recent series of Amazon Alexa Prize Challenges providing a competition platform and monetary incentives to spur development~\citep{Ram:2017, DBLP:journals/corr/abs-1812-10757}. 
Many practical applications of conversational agents have been proposed, e.g., for companionship to improve mental well-being, (e.g., \citep{morris2018towards}) and for therapy (e.g., \citep{fitzpatrick2017delivering}). Open-domain conversational agents can also be used in educational settings as tutors and evaluators, as proposed in~\citep{fonte2009tq}. 
While much room for improvement remains in the current implementations of the conversational AI systems, the potential for intelligent, empathic, and broad-coverage conversational agents is widely recognized.

However, for an open-domain conversational agent to be coherent and engaging, it must be able to drive the conversation to the next topic, and in a way that does not appear scripted. This task is complicated. As for many realistic and complex tasks, extensive knowledge engineering is needed for in-depth domain-specific capabilities, usually handled by specialized components. 
For a user to remain engaged, the overall conversational AI system should be able to recommend the next conversation topic (or component) in a natural and coherent fashion. Appropriate topic recommendations are also critical to expose the capabilities of the system to the user, who otherwise may not know that a conversational agent is an expert in particular topics like sports, cars, or video games.

Yet, the right topic to recommend depends on both prior user interests and the conversation context. Extensive work has been done in topic and content recommendation using content-based \citep{lops2011content} and collaborative filtering methods \citep{mnih2008probabilistic,koren2009matrix}. However, it is unclear just {\em how} to adapt recommender system techniques to the conversational setting. In open-domain conversational agents, the recommended items might be agent's sub-components with dynamically changing content and interactions underneath, specific items, general suggestions, or even clarification questions. In this paper, we extend model-based and collaborative filtering recommendation algorithms for topic and content recommendation in the conversational setting. Our contributions are threefold: (1) Formalizing the conversational topic suggestion problem for open-domain conversational agents; (2) Development of a sequential contextual topic suggestion model for this task; (3) Empirical exploration of the effectiveness of model-based, collaborative filtering-based, and hybrid approaches to topic suggestions in the conversational setting. Our experiments indicate the value of our proposed hybrid recommendation solution, highlight the challenges and opportunities inherent in the conversational topic recommendation, and suggest promising directions for future work. 

\begin{table*}[h!]
    \fontsize{7}{7}
    \selectfont
    
    \begin{tabular}{ p{0.5cm} |p{4.1cm} | p{6.7cm} | p{0.7cm} | p{0.78cm} | p{0.7cm} | p{1.5cm}}
    \bottomrule
    \textbf{\small Turn} & \textbf{\small User Utterance} & \textbf{\small System Response}  & \textbf{\small Topic} & \textbf{\small PS} & \textbf{\small PST} & \textbf{\small Labels} \\
    \bottomrule
    \bf 1&Alexa, let's chat.  & Hi! What would you like to talk about?  & \textit{Music} & None & {\em None} & $chat$\\
    \bf 2&Tell me recent songs. & Circles by Post Malone, Talk by Khalid, and Bad Guy by Billy Eilish are currently popular. Would you like to talk about Billy?  & \textit{Music} & \textit{Opening} & {\em Music}& $Music\_{accept}$ \\
    \bf 3&No I do not. & Alright. How about some info on Khalid? & \textit{Music} & \textit{Music} & \textit{Music} & $follow-up$\\ 
    \bf4& Oh, no. & Ok, do you want to hear about some places to travel? & \textit{Phatic} & \textit{Music} & \textit{Music} & $chat$\\
    \bf 5&I love traveling. & Cool! Which country do you want to visit? & \textit{Travel} & \textit{Music} & \textit{Travel} &  $Travel\_{accept}$ \\
    \bf 6&Somewhere in Australia.  & Australia is beautiful, especially the beaches. Do you like beaches? & \textit{Travel} & \textit{Travel} & \textit{Travel} & $follow-up$ \\
    \bf 7& Yes. & I recommend Bondi Beach, it is amazing. Wanna here more about it? & \textit{Travel} & \textit{Travel} & \textit{Travel} & $follow-up$ \\
    \bf 8&No thanks, let's talk about something else. & Sure, would you like to discuss the news? & \textit{Phatic} & \textit{Travel} & \textit{Travel} & $chat$ \\
    \bf 9&No, news is boring. & Sure, Do you want to talk about recent movies? & \textit{Phatic} & \textit{Phatic} & \textit{News} & $News\_{reject}$ \\
    \bf 10&Okay, that sounds interesting. & Cool, which genre do you prefer, comedy or drama? & \textit{Movies} & \textit{Phatic} & \textit{News} & $Movie\_{accept}$\\
    \bf 11&I like both. & How about The Favourite by Yorgos Lanthimos? & \textit{Movies} & \textit{Movies} & \textit{Movies} & $follow-up$\\
    \bf 12& I have to go Alexa, bye! & Nice talking to you, good bye! & \textit{Phatic} & \textit{Movies} & \textit{Movies} & $chat$
     \\
    \bottomrule
    \end{tabular}
    \caption{A Conversation example, where PS and PST refer to Previous State and Previous Suggested Topic, respectively. }
\label{tab:conversationsample}
\vspace{-1cm}
\end {table*}

\vspace{-0.5cm}
\section{RELATED WORK}
\label{sec:related}
To put our work in context, we briefly review the published research on open-domain conversational agents, key techniques used for general recommender systems, utterance suggestion approaches used in conversational agents, and recent research on the single-domain conversational recommendation.

\vspace{-0.2cm}
\subsubsection*{\bf General chatbots and conversational agents. }
Recent years witnessed significant research activity in developing a coherent and engaging conversational agent \citep{wen2016network, Thomas:2018, Jason:2017, Xuesong:2017}. conversational agents have been generally categorized ~\citep{Su:2018} into two main categories, namely task-oriented and general chat. Chatbots are traditionally aimed primarily at \textit{small talk}, while task-oriented models are designed to carry out information-oriented and transactional tasks ~\citep{Hang:2016, Radlinski:2017}. Recently, several shared tasks and challenges have been proposed to push the boundaries of conversational AI to develop more intelligent chatbots to carry on in-depth conversations about a number of topics, not just \textit{small talk}. This research has been evaluated by both crowd workers~\citep{DBLP:journals/corr/abs-1811-01241} and live users as part of the Alexa Prize Conversational AI challenge ~\citep{Ram:2017, DBLP:journals/corr/abs-1812-10757}. 

\subsubsection{\bf General Recommender Systems Research}
Recommender systems have been studied for decades, and are now pervasive \citep{ricci2011introduction}. Traditional recommendation algorithms have been classified as primarily {\em model-based} or {\em content-based}, where a classifier model is trained for each user's profile, and {\em collaborative filtering}, where a user's unknown preferences are estimated based on the neighborhood of similar users \citep{herlocker1999algorithmic}. More effective methods have been shown to be a hybrid of the two approaches \citep{melville2002cbcf}, with increasingly sophisticated methods reported for movie \citep{koren2008factorization, koren2009matrix} and news recommendation \citep{das2007google}. However, in all cases, there are significant challenges that still remain active topics for research. The most closely related issue is the {\em cold start} problem, i.e., recommending an item for a new user with no existing profile, or recommending new (or dynamic or changing) items, with no prior likes from any users. Model-based or content-based recommendations have been shown to perform better in such scenarios\citep{wang2018dkn, lu2015content}. This is the primary approach we attempt to adopt here for the conversational topic recommendation. Other approaches have explored online experimentation (e.g., ~\citep{li2010contextual}), and using social media or other metadata (e.g., ~\citep{chen2010short,phelan2009using} for recommendation. Unfortunately, these signals are not easily available in the conversational setting.
All the attributes of users and their preferences need to be inferred from their interaction with the conversational agent.

\subsubsection{\bf Utterance Suggestion in conversational agents}
Yan et al. \citep{yan2018smarter} describe an end-to-end generative model, which given a user query, generates a response, and a proactive suggestion to continue the conversation.
However, generative models like this still strictly rely on training corpora or restricted information, without the ability to query external data sources, thus limiting their capacity for an informative conversation. 
In the other work, Yan et al. ~\citep{yan2017joint} describe a next-utterance suggestion approach for retrieving utterances from a conversational dataset to use as suggestions, along with the response. The proposed model learns to give suggestions related to the response, to continue the conversation on the same topic.
In practice, due to the vast number of possible utterances coming into a social bot, many conversational systems rely on multiple response modules where each response module would be responsible for a particular domain or set of domains ~\citep{khatri2018advancing}.
Fine-grained utterance suggestions would be applicable to the implementation of each domain-specific module. However, when the user is passive or gets fatigued with a particular topic, the system needs to switch to a different component with domain-specific capabilities to keep the user engaged. In this paper, we attempt to formalize the problem of suggesting the best next interesting topic. 

\vspace{-0.2cm}
\subsubsection*{\bf Conversational Recommendation. }
Recently, the idea of conversational recommendation was introduced~\cite{christakopoulou2016towards}, primarily as a way to elicit the user's interests for item recommendation. For example, Sun at al. \citep{sun2018conversational} introduced an end-to-end reinforcement learning framework for a personalized conversational sales bot, and in \citep{li2018towards}, a combination of deep learning-based models is used for conversational movie recommendation. Currently, the most existing conversational agents are designed for a single domain, such as \textit{Movies} or \textit{Music}. An open-domain conversational agent that coherently and engagingly converses with humans on a variety of {\em topics}, remains an aspirational goal for dialogue systems ~\citep{Venkatesh:2017, Ram:2017}. To address this problem, we propose a conversational {\em topic} suggestion method for open-domain conversational agents in which the conversational agent proactively suggests the next best topic to discuss based on the conversation so far, as we describe next.

\begin{table*}
\begin{center}
\fontsize{8pt}{8pt}
\selectfont
    \begin{tabular}{l|l|l}
    \hline
    {\bf \em  \large {Features}} & {\bf \em \large Description} & {\bf \em \large Example Values}\\
    \hline \hline
    \bf Topic and Behavior &  &  \\
    \textit{F}\textsubscript{1} -- \textit{F}\textsubscript{8} & 1-hot encoding for user response for each topic on previous turn, & \\
    & where 1=Accepted, 0=not suggested, and -1=rejected & [+1,0,-1,0,0,0,0,0]\\
    \textit{F}\textsubscript{9} -- \textit{F}\textsubscript{10} - Two previous topics & Two previous components that user engaged with& Movies, Music\\
    \textit{F}\textsubscript{11} -- Previous accepted topic & Previous suggestion that was accepted by the user& Music\\
    \textit{F}\textsubscript{12} -- Previous rejected topic & Previous topic that was rejected by users & Pets\_Animals\\
    \hline
    \hline
    \bf User Profile &  & \\ 
    \textit{F}\textsubscript{13} -- Name & Does user give his/her name & True/False\\
    \textit{F}\textsubscript{14} -- Gender & What is the user's gender & M/F\\
    \textit{F}\textsubscript{15} -- Time & Time of the day during the conversation & Morning/Day/Evening/Night\\
    \hline
    \end{tabular}
\end{center}
\caption{Dialogue manager state information features used for CTS recommendation. The values are computed up to turn $i$ in the conversation so far.  }
\label{tab:Features}
\vspace{-1cm}
\end {table*}
\vspace{-0.3cm}
\section{CONVERSATIONAL TOPIC SUGGESTION (CTS): PROBLEM DEFINITION}


We now define the conversational topic suggestion problem and introduce our proposed solutions in the following section. 

Consider the example conversation in Table \ref{tab:conversationsample}. While this is not a real user\footnote{Exact user conversations cannot be reproduced due to Alexa Prize terms.}, the conversation is typical of those observed with our system during the Alexa Prize challenge. In a regular Alexa conversation, a user may have an initial interest or information need (e.g., ``recent songs'') which is handled by a particular system component (in this case, the {\em Music} component); however, the user might quickly lose his/her interest, and the system (conversational agent) must take the initiative to find the next topic of conversation that this user is likely to be interested in, for example, {\em Travel}.
In the example conversation, the user accepts the suggestion to talk about the topic {\em Travel}, and a different system component starts interacting with the user to drive the conversation. The next suggested topic {\em News}, however, is not accepted by the user, and the system has to make another recommendation, which would degrade the user experience.

We define Conversational Topic Suggestion (CTS) as follows:

\vspace{-0.3cm}
\begin{figure}[h!]

\centering
\noindent\begin{tabular}{l l}
\hline \hline
 \textbf{Setting:} & Open-Domain mixed initiative conversation with a\\
                    & multi-component conversational agent. \\
 \textbf{Given:} & A conversation $C$, consisting of a sequence of user \\
                & utterances $U_{0..i}$, a sequence of system states $S_{0..i}$, \\
                & and a set of possible conversation topics $t \in T$, \\
                & (e.g., system components or {\em \bf mini-skills}). \\
 \textbf{Problem:} & At conversation turn $i$, select a topic $t_i$ to suggest\\
                    & for the current user $u$, to maximize the likelihood\\
                    & of {\em acceptance} (i.e., the probability that user $u$\\
                    & would like to talk about the topic $t_i$ next). \\
 \hline
 \\
\end{tabular}
\label{fig:cts-problem}
\vspace{-0.8cm}
\caption*{Definition 1: Conversational Topic Suggestion (CTS) Problem Statement.}
\end{figure}
\vspace{-0.3cm}

Note that this definition focuses on the acceptance of the topic suggestion, and does not explicitly consider the user's future satisfaction or engagement with the selected topic (e.g., as measured in reference~\cite{choi2019offline}). We plan to explore the connection of topic suggestion and ultimate user satisfaction with the conversation in future work. We also emphasize that we formulate CTS based only on short-term history (conversation or session-level), and not on long-term user interests. Long-term personalization is also a promising direction for future work. Still, it is worth noting that in many practical situations, a conversational agent must make coherent topic suggestions for new (cold-start) or inactive users. Secondly, note that we do not require a set of other users (e.g., as would be required for collaborative filtering approaches for recommendation), which might be an attractive setting for privacy and security considerations. Finally, unlike in a traditional recommendation setting, a suggested topic represents a system component which, if accepted, begins interacting with the user dynamically, and thus cannot be easily mapped to a single {\em item} that can be easily represented and compared across users. Our models, described next, attempt to capture both the conversation context and the internal system information for this task.

\vspace{-0.2cm}
\section{CTS-Seq APPROACH}
\label{cts-seq-approach}
For relevant and coherent topic suggestions, it is necessary to consider the conversation context, e.g., the sequence of previous user utterances and system states. For example, if a user is talking about \textit{Movies}, it might be more natural to suggest \textit{Music} as the next topic, as opposed to \textit{Cars}. Also, if a user declined to talk about \textit{Movies} in the past, the system should not suggest this topic or a related topic like \textit{Television} unless explicitly requested. For this reason, we propose to use a sequence modeling approach for the conversational topic suggestion. We will further show how this approach can be combined with more traditional collaborative filtering-based methods.

Before we describe the specific sequence models, we first discuss the conversation {\em features}, used for both sequence modeling and collaborative filtering-based methods. 

\vspace{-0.3cm}
\subsection{\bf System State and User Profile Features}
\label{sec:features}

To represent the conversation context, two different groups of features are extracted for each conversation turn, as summarized in Table~\ref{tab:Features}. The first group is {\em Topic and Behavior} features, which represents the user's previous responses, i.e., the accepted and rejected topic suggestions. 
These features have the values of $\mathbf{1}$ for accepted topics, $\mathbf{-1}$ for rejected topics, and $\mathbf{0}$ for the topics that have not been proposed yet. 
This group of features is designed to prioritize the topics that have been accepted $\mathbf{1}$ or unexplored $\mathbf{0}$. {\em Topic and Behavior} features also model topic classification features and the current conversation context and system state. These features could indicate the historical probability that the current state is a potential topic-switching point, or whether it should be a {\em follow-up} for the previous topic. The second group of features is {\em User Profile} features. They contain the inferred gender of the user [-1,1] based on the provided name, and whether they gave their name at the start of the conversation or not (a weak indicator of the user's openness to sharing information with the bot). Other features like age and location, which are often used for user profiling, are usually not available in the conversational setting. Table \ref{tab:Features} shows different categories of features that are used in all the CTS-CRF, CTS-CNN, and CTS-RNN models. The values of these feature groups are computed for each conversation turn and stored in separate vectors, which are then concatenated to produce the full conversational state representation, $fv$, specifically: 
\vspace{-0.3cm}

\begin{equation}
     fv = [F\textsubscript{1}; F\textsubscript{2};...; F\textsubscript{15}] 
\end{equation}
\vspace{-0.3cm}

We emphasize that these features will be used for all sequence and CF-based model variations, to explore the trade-offs in modeling, while keeping the actual features constant. 

\vspace{-0.2cm}
\section{CTS-Seq: MODELS}
\label{cts-seq-models}
In this section, we list three different implementations of the proposed CTS-Seq method.
First, we describe the Conditional Random Fields (CRF) implementation. Then, we describe the CNN-based followed by the RNN-based implementation.

\subsection{CRF Implementation of CTS-Seq: CTS-CRF}
\label{cts:crf}

As the first and most straightforward implementation of CTS, we use the well-known and robust CRF model. CRF is an undirected graphical model, which estimates the conditional probability of a sequence of labels (tags) with respect to the observed features, and requires relatively small amounts of training data~\citep{crf2001Lafferty,crf2018Ye}.

Each conversation is represented as a sequence of turns, with observable features extracted from each utterance and system state.  
Recall that we represent a conversation $j$ as a sequence of turns $Conv_{(j)} = [utt_{(1)},...,utt_{(i)},..., utt_{(n)}]$. Then, for each sequence of utterances, a sequence of labels (topics) $[t_1,...,t_i,...,t_n]$ is generated. The recommended topic $t$ is modeled as the CRF hidden state, and $X$ is the observed variable represented by the features described above. Thus, the CTS-CRF model aims to predict the most likely {\em next} topic $t_{(i+1)}$ after observing the first $i$ conversation turns and system states.

More formally, Eq. \ref{eq:crf} computes the probability of a topic $t$ given the sequence of previous turns and topic decisions, where $Z(X)$ indicates the normalization factor and $\theta$ and $\eta$ are weights that can be tuned using maximum likelihood estimation. Moreover, $f(t_i;X_t)$ and $g(t_i;t_{i-1};X_t)$ jointly represent the next topic to predict, the context (previous topic) and the features for the current turn $x$.

\begin{equation}
\begin{split}
p(t|x) \propto \frac{1}{Z(X)} \prod_{i=1}^m \exp^{\Big\{ \sum_{j=1}^m {\theta_j f_j (t_i; X_t)} + \sum_{k=1}^m {\eta_k g_k (t_i;t_{i-1}; X_t)} \Big\} }
\end{split}
\label{eq:crf}
\end{equation}

\begin{figure}[htb]
\includegraphics[width=200 pt]{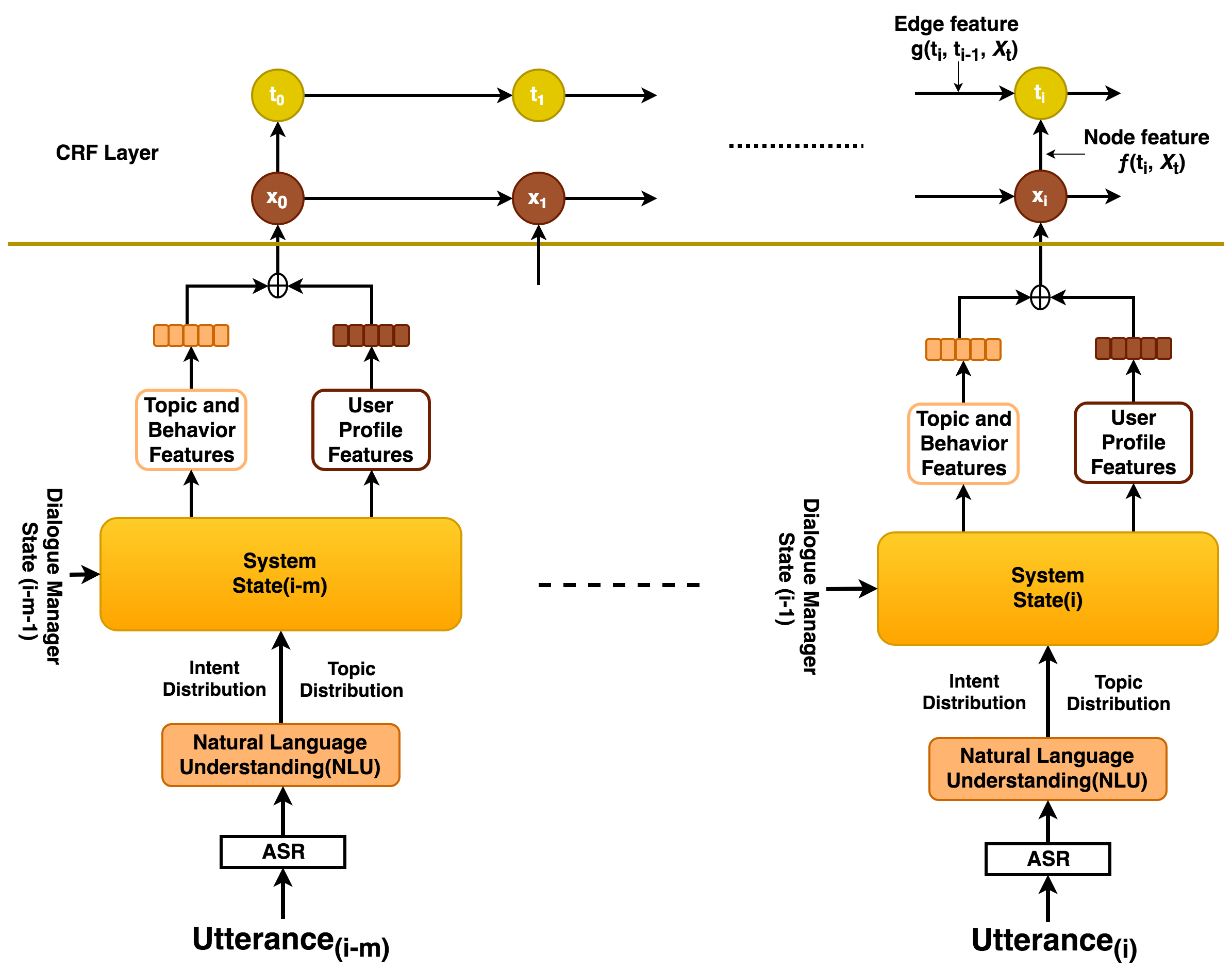}
\caption{CTS-CRF topic suggestion model for conversation turn $i$. Feature details are reported in Table~\ref{tab:Features} and Section~\ref{sec:features}. ASR stands for authomatic speech recognition.}
\label{fig:CRF}
\vspace{-0.7cm}
\end{figure}

The CRF-based implementation of CTS, CTS-CRF, is illustrated in Figure~\ref{fig:CRF}.

\begin{figure*}[h]
\centering
\includegraphics[width=400 pt]{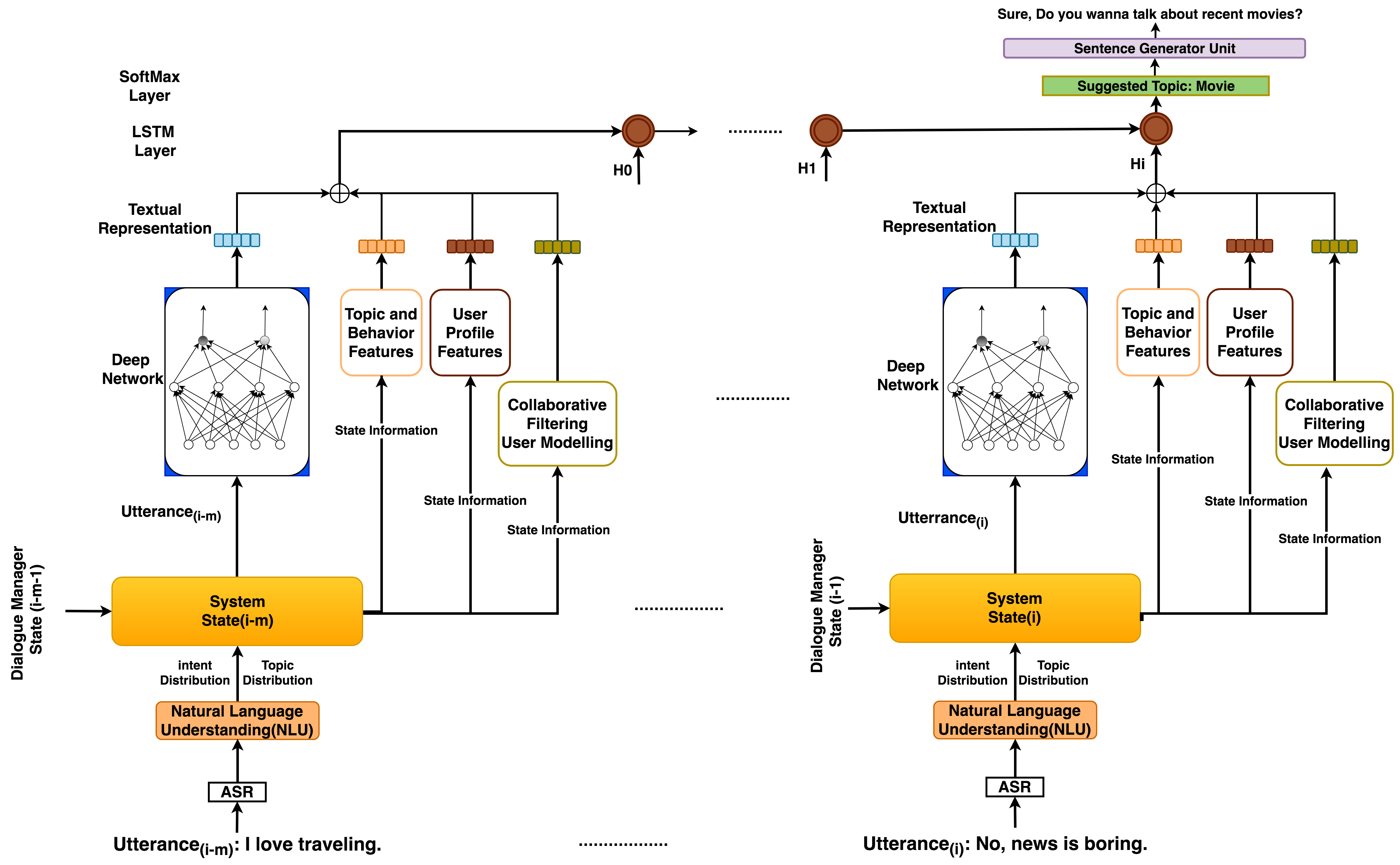}
\caption{CTS-RNN-CF or CTS-CNN-CF model architecture, where \textit{Topic and Behavior} features include the list of all previously suggested, accepted and rejected topics from the beginning of the conversation. \textit{User Profile} features contain the list of Name, predicted Gender, and time of the day. CF features also include the suggested topic distribution extracted from the collaborative filtering model. Feature details are reported in Table~\ref{tab:Features}. ASR stands for automatic speech recognition.}
\label{fig:CTS-RNN}
\vspace{-0.3cm}
\end{figure*}

\subsection{Deep-learning based implementation of CTS-Seq: CTS-CNN and CTS-RNN}
\label{cts:cnn}

Deep learning approaches such as Convolutional Neural Networks (CNNs) and Recurrent Neural Networks (RNNs) have shown promising results for different natural language processing tasks, from text classification to dialogue act classification (e.g., ~\citep{Bengio:NLM, lecun:scratch, Conneau:VDNN, cer2018Daniel, kim:2014, DA-Blunsom2016, Lee:seq}). Lee et al. \citep{Lee:seq} proposed a pipeline of deep learning methods to model a {\em sequence} of short texts. Inspired by \citep{Lee:seq}, we propose two deep learning models for implementing CTS, namely CTS-CNN and CTS-RNN. CTS-CNN and CTS-RNN respectively use a CNN and a BiLSTM to incorporate textual and contextual evidence gathered so far from the conversation to recommend (predict) the next conversation topic.

\subsubsection*{\bf CTS-CNN Implementation. }

Here, we walk through different steps in the CTS-CNN model.
CTS-CNN-CF network takes word tokens, from $m$ consecutive utterances. $utt_{(i)}$ stands for the words in the $i$-th utterance, where $w_{ij}$ stands for $j$-th word in $i$-th utterance.
\begin{gather}
utt_{(i-m)} = [w_{(i-m)1}; w_{(i-m)2}; w_{(i-m)3} \ ... \ w_{(i-m)n}] 
\label{eq:words1}
\end{gather}
\hspace{125 pt} ....
\begin{gather}
utt_{(i-1)} = [w_{(i-1)1}; w_{(i-1)2}; w_{(i-1)3} \ ... \ w_{(i-1)n}] 
\\
utt_{(i)} = [w_{i1}; w_{i2}; w_{i3} \ ... \ w_{in}] 
\label{eq:words2}
\end{gather}
We define a function $f_c$ that takes an utterance as input and outputs the learned utterance representation $y_{cnn}$:
\begin{gather}
y_{(cnn_{(i)})} = f_c(utt_{(i)}) 
\end{gather}
$f_c$ is a 3-layered CNN with max pooling, which is applied in parallel on all the utterances in a window of size $m$. The first layer is a word embedding layer with pre-initialized weights from Word2Vec vectors of size 300. The weights on the embedding layer are tuned during training using the cross-entropy loss function.

\subsubsection*{\bf CTS-RNN Implementation. }

CTS-RNN uses a BiLSTM network followed by an attention layer to model the utterance representation. In CTS-RNN, a function $f_r$ is defined that takes an utterance as input and outputs a hidden representation $h_{(i)}$ for each utterance:

\begin{equation}
   h_{(i)} = f_r(utt_{(i)})  
\end{equation}

where $f_r$ is a BiLSTM model with 256 hidden layers. It is also applied in parallel on a window of size $m$ in the same way as for $f_c$. Then, the hidden representation for the $i$-th utterance is passed to an attention layer to generate the final representation $y_{(rnn_{(i)})}$. Given the hidden representations of each timestamp of $j$ in $LSTM_{(i)}$ is $h_{ij}$, dot product similarity score $s_{ij}$ is computed based on a shared trainable matrix ${M_i}$, context vector $c_i$ and a bias term $b_{ij}$. ${M_i}$, $c_i$ and $b_{ij}$ are initialized randomly and jointly learned during training.  Softmax activation is applied on similarity scores to obtain attention weights $\alpha_{ij}$. Lastly, using learned $ \alpha_{ij} $, a weighted sum on BiLSTM hidden representations is applied to obtain the output $y_{rnn_{(i)}}$ for the $i$-th utterance as follows:

\begin{equation}
 s_{ij} = tanh \left((M_{i})\textsuperscript{T}{h}_{ij} + b_{ij} \right )
\end{equation}

\begin{gather}
    \alpha_{ij} = \frac{e^{(s\textsubscript{ij})\textsuperscript{T}c_{i}}}{\sum_{j=1}^{n}e^{(s\textsubscript{ij})\textsuperscript{T}c_i} }
    \to
    y_{(rnn_{(i)})} = \sum_{j=1}^{n} \alpha\textsubscript{j} h\textsubscript{ij} 
\end{gather}

Finally $y_{(rnn_{(i)})}$ is computed for every utterance located in the window.

\subsubsection*{\bf Merging and FeedForward Layers. }
\label{merging-layer}

This step is similar for both CTS-CNN and CTS-RNN models, where the output of the textual representation of each utterance is merged with {\em Topic and Behavior}, and {\em User Profile} features. Here we describe all the details of these layers for the CTS-CNN model.

To create the final representation of a $w_{(i)}$ in a conversation, we extract $y_{(cnn_{(i)})}$ from all the utterances located in the window in parallel. Then, the window is fed to an LSTM network with 100 hidden states. We deploy an LSTM instead of an Bi-LSTM since in real conversation, there is not a backward signal. Finally, the output of the last layer going through 
\begin{gather}
w_{(i)} = [utt_{(i-m)};...;utt_{(i-m+j)};...;utt_{(i)}]\\
rep_{(w_{(i)})} = [y_{(cnn_{(i-m)})};...;y_{(cnn_{(i-m+j)})};...;y_{(cnn_{(i)})}] \\
output = LSTM \Big ( rep_{(w_{(i)})} \Big)
\label{eq:window}
\end{gather}

Where, in this equation $j < m < i$. The final output is fed to a feed forward layer of size 256 with a dropout rate of 0.5. A softmax function $f(s)$ is applied to generate a probability distribution over $C$ possible topics. The network was trained with an Adam optimizer with a learning rate of 0.001 using the softmax cross-entropy loss function $CE$. $C$ is the number of classes, $t_i$ is the one-hot representation of the target label, and $s_i$ are the scores inferred by the model for the $i$-th class:

\vspace{-0.5cm}
\begin{gather}
    f(s_i) = \frac{e^{s_i}}{\sum_{i=j}^{C} e^{s_j}} \to CE = - \sum_{i=1}^{C} t_{i} \log \left ( f  (output)  )\right ) 
\end{gather}

We summarize the parameters of the (CNN-) and (RNN-) based models in Table \ref{tab:par}. The parameters were not tuned and were chosen based on our experience and previous literature. 

\begin{table}
\selectfont
  \begin{tabular}{cc}
    \toprule
    Parameters & Values\\
    \midrule
    Pooling type  & Max-pooling\\
    L2-regularization  & 0.001 \\
    Word embedding length & 300 \\
    Momentum & 0.997\\
    Epsilon & 1e-5 \\
    Learning rate & 0.001 \\
    Dropout & 0.5 \\
    Batch Size & 64\\
    Number of layers (CNN)  & 3 \\
    Number of filters(CNN) & 128 \\
    Filter sizes (CNN) & 1,2,3 \\
    Hidden state (RNN) & 100 \\
    Feed forward layer (RNN) & 256 \\
    
 \bottomrule
\end{tabular}
\caption{Detailed configuration parameters of the CTS-CNN and CTS-RNN implementations.}
\label{tab:par}
\vspace{-0.9cm}
\end{table}


\subsection{Hybrid Sequential and Collaborative Filtering: CTS-Seq-CF}
\label{sec:hybrid-model}

We now introduce our new model, CTS-Seq-CF, which augments the sequence modeling approach described above, with additional signals extracted from other users' experiences using collaborative filtering. The proposed models, CTS-CNN-CF and CTS-RNN-CF, incorporate the probability of acceptance of each topic based on similar users' preferences (We describe the collaborative filtering methods used in Section \ref{sec:personalized-suggestion}.) as features into the CTS-CNN and CTS-RNN models, respectively. One of the resulting hybrid models, CTS-RNN-CF, is illustrated in Figure \ref{fig:CTS-RNN}. CTS-CNN-CF follows the same pattern as CTS-RNN-CF, with the semantic utterance representation being generated using a CNN model. They aggregate contextual evidence from the preceding states by considering a window of size $m$ for each turn. Then, all the system state information for the turns in that window is extracted, which includes all the features in Table \ref{tab:Features}, as well as suggested topic distribution predicted by the collaborative filtering method described in Section \ref{sec:personalized-suggestion}. Finally, all the utterance vector embeddings within the window are concatenated with them to form the window vector embedding. 
Here, we walk through the details for CTS-RNN-CF. 

$y_{(rnn_{(i)})}$ is first concatenated with $fv_{(i)}$ to obtain the enriched representation $rep_{(rnn_{(i)}+fv_{(i)})}$ of an utterance. Then, we concatenate them with the CF features, which were extracted by collaborative filtering module to generate the final utterance representation $rep_{utt_{(i)}}$.

\vspace{-0.5cm}
\begin{gather}
rep_{(cnn_{(i)} + fv_{(i)})} = [y_{(cnn_{(i)})};fv_{(i)}] 
\\
rep_{(utt_{(i)})} = [rep_{(cnn_{(i)} + fv_{(i)})};CF_{(i)}] 
\end{gather}

To create the final representation of a $w_i$ in a conversation, we extract $rp_{(utt_{(i)})}$ from all the utterances located in the window in parallel. Finally, all the outputs are concatenated together to form the final vector.
\begin{gather}
w_{(i)} = [utt_{(i-m)};...;utt_{(i-m+j)};...;utt_{(i)}]\\
rep_{(w_{(i)})} = [rep_{(utt_{(i-m)})};...;rep_{(utt_{(i-m+j)})};...;rep_{(utt_{(i)})}]\\
output = LSTM \Big ( rep_{(w_{(i)})} \Big)
\label{eq:window:cts-seq-cf}
\end{gather}

Then, $rep_{(w_{(i)})}$ is fed to an LSTM network with 100 hidden states, later the output of the last layer going through a feed-forward layer followed by a softmax layer as described in Section \ref{merging-layer}. 

\vspace{-0.2cm}
\section{EXPERIMENTAL SETUP}

We now describe the baselines, data, metrics, and experimental procedures used to evaluate our proposed conversational topic suggestion models.
\vspace{-0.2cm}
\subsection{Baseline 1: Popularity Method}
\label{sec:popularity}

The Popularity method is a heuristic method, which suggests the next conversation topic based on overall frequency (popularity) in previous conversation data and previous user ratings, and the approximate time of day. The order of suggestion is {\em Movies}, followed by {\em Music}, followed by {\em Video Games} or {\em Travel} or {\em Animals} depending on the time of day to accommodate the expected differences in user demographics. This heuristic popularity baseline was deployed during the Alexa Prize competition \cite{ahmadvand2018emory}.


\vspace{-0.2cm}
\subsection{Baseline 2: Collaborative Filtering (CF)}
\label{sec:personalized-suggestion}



We adapt the classical approach of CF, originally introduced for item recommendation to the conversational setting using the K-Nearest Neighbors (KNN) model. Each user is represented by the following features:
\begin{itemize}
    \item User and time features: $F_{13}$,  $F_{14}$,  $F_{15}$ from Table \ref{tab:Features}.
    \item Suggestion acceptance and rejection rate: The fraction of topic suggestions accepted by the user and the fraction rejected by the user.
    \item Topical features: $F_1$ - $F_8$ from Table \ref{tab:Features}.
\end{itemize}

For each conversation turn, the feature vector described above is calculated based on the conversation up to this turn. For example, if a user has accepted a suggestion to talk about \textit{Movies} and rejected a suggestion for \textit{Music}, the accept and reject rates would be 0.5. The topical feature vector would contain $\mathbf{1}$ for \textit{Movies} and $\mathbf{-1}$ for \textit{Music}, and then top $k$ users with most similar conversation histories would be retrieved.
More formally,
\begin{equation}
  \mathbf{U}_a = [F_1(a):F_8(a),F_{13}(a):F_{15}(a),r^{accept}{(a)},r^{reject}{(a)}]  
\end{equation}
\begin{equation}
        sim(\mathbf{U}_a,\mathbf{U}_b) = \frac{\mathbf{U}_a \cdot \mathbf{U}_b}{ ||{\mathbf{U}_a} ||\times||{\mathbf{U}_b}||} 
\end{equation}
\begin{equation}
pred(\mathbf{U}_a,T) =\frac{\sum_{\mathbf{U}_b \in N} sim(\mathbf{U}_a,\mathbf{U}_b) \times s_{(\mathbf{U}_b,T)}} {\sum_{\mathbf{U}_b \in N} sim(\mathbf{U}_a,\mathbf{U}_b)}
\end{equation}

where $\mathbf{U}_a$ is the user who we are calculating the topic scores for, $\mathbf{U}_b$ is one of the neighbors from set $N$, which is the set of 33 nearest neighbors of $\mathbf{U}_a$,  $r^{accept}{(a)}$ is the suggestion acceptance rate of user $\mathbf{U}_a$, $r^{reject}(a)$ is the suggestion rejection rate of user $\mathbf{U}_a$, $s_{(\mathbf{U}_b,T)}$ indicates the score of topic $T$ for user $\mathbf{U}_b$,
 and $pred(\mathbf{U}_a,T)$ represents the predicted score of a topic $T$ for the active user.
 
For final classification, the predicted topic scores based on 33 nearest neighbors' preferences are fed to a feed-forward layer followed by a softmax layer, as described in Section \ref{merging-layer}.

\vspace{-0.2cm}
\subsection{Baseline 3: Contextual Collaborative Filtering: Contextual-CF}
\label{sec:contextual-CF}
Contextual-CF utilizes the collaborative filtering signals extracted from the preceding utterances. Then, a fully connected neural network followed by a softmax is applied to combine the features and provide the final prediction result. To this end, we applied the CF model described in Section \ref{sec:personalized-suggestion} to extract the suggested topics for all the utterances located in a window of size $m$. To represent the CF features, we considered a one-hot-vector, where the length of the one-hot-vector is equal to the number of available topics that are supported by the conversational agent. The value corresponding to the topic selected by the CF model is assigned as $\mathbf{1}$. Then, the one-hot-vectors are concatenated together to create the final vector for the corresponding window. As a result, a vector of size $[window\_size * {len(one-hot-vector)]}$ is generated. Eq. \ref{contextual-cf} represents the feature vector of $w_{(i)}$.
\begin{equation}
w_{(i)} = [utt_{(i-m)};...;utt_{(i-m+j)};...;utt_{(i)}]\\
\end{equation}
\begin{equation}
CCF_{(w_{(i)})} = [CF_{(utt_{(i-m)})};...;CF_{(utt_{(i-m+j)})};...;CF_{(utt_{(i)})}]
\label{contextual-cf}
\end{equation}

Where $CCF_{(w_{(i)})}$ indicates the contextual CF features extracted from $i$-th window and $CF_{(utt_{(i)})}$ represents the CF features extracted from the $i$-th utterance. For final classification, $CCF_{(w_{(i)})}$ is fed to a feed forward layer followed by a softmax layer as described in Section \ref{merging-layer}.

\vspace{-0.2cm}
\subsection{Methods Compared }

For convenience, we summarize the methods compared in the next section for reporting the experimental results.

{\bf Popularity}:A heuristic method, described in Section~\ref{sec:popularity}, using topic frequency in previous conversations.

{\bf CF}: The collaborative filtering approach, described in Section~\ref{sec:personalized-suggestion}, using the conversation state (accepted/rejected topic suggestions) as the user profile.

{\bf Contextual-CF}: The contextual collaborative filtering approach, described in Section~\ref{sec:contextual-CF}, incorporating CF signals from preceding utterances into CF features from the current utterance using a fully connected neural network.


{\bf CTS-CRF}: The CRF implementation of the CTS approach, described in Section~\ref{cts:crf}, using only the conversational context (model-based recommendation).

{\bf CTS-CNN}: The CNN implementation of the CTS approach, presented in Section~\ref{cts:cnn}, using only the conversational context features (model-based recommendation).

{\bf CTS-RNN}: The RNN implementation of the CTS approach, presented in Section~\ref{cts:cnn}, using only the conversational context features (model-based recommendation).


{\bf CTS-CRF-CF}: The hybrid model-based and collaborative-filtering based approach, enhancing the CTS-CRF model with collaborative filtering features (Section~\ref{cts:crf}).

{\bf CTS-CNN-CF}: The hybrid model-based and collaborative-filtering based approach, enhancing the CTS-CNN model with collaborative filtering features (Section~\ref{cts:cnn}).

{\bf CTS-RNN-CF}: The hybrid model-based and collaborative-filtering based approach, enhancing the CTS-RNN model with collaborative filtering features (Section~\ref{cts:cnn}).

\vspace{-0.3cm}
\subsection{ Dataset: Amazon Alexa Prize 2018}
The conversation data were collected by participating in Amazon Alexa Prize 2018 competition~\citep{DBLP:journals/corr/abs-1812-10757}. The conversation dataset consisted of 14,707 open-ended conversations longer than four turns (because the first 2-3 turns usually consisted of the required introduction and exchanges of greetings). These conversations were collected from August 1, 2018, to August 15, 2018. The first ten days of conversations were used for training and the rest for testing. The relative topic popularity is shown in Table \ref{tab:alexadist}. The conversations have an average length of 11.5 turns, where 91\% of the conversations contain at least one suggestion, and 60\% have at least two explicit topic suggestions.
\begin{table}[htbp]
\footnotesize
      \centering
          \begin{tabular}{l l|l l|l l}
           \toprule
          Movie &20.1\%&Music&14.4\%& News&18.4\%  \\ 
          Pets\_Animal&10\%&Sci\_Tech&6\%&Sports&6\% \\ 
          Travel&9.1\%& Games&6\%&Celebrities&2.5\% \\ 
          Literature&1.5\%&Food\_Drinks&1.5\% &Other&1.5\% \\
          Weather&1.5\%&Fashion&1\%&Fitness&1\%  \\
          Entertainment and Cars&1\%&&\\
          \bottomrule
         \end{tabular}
      \caption{Topics distribution in Alexa 
      dataset.}
      \label{tab:alexadist}
\vspace{-1cm}
\end{table}

\vspace{-0.2cm}
\subsection{Evaluation Metrics}

\begin{table*}[htb]
\centering
\small
\begin{tabular}{l|c|c|c|c|c|c|c|c|c}
\bottomrule
 Suggested Topic / Method                & Popularity & CF & C-CF &  CTS-CRF & CTS-CNN & CTS-RNN & CTS-CRF-CF & CTS-CNN-CF &CTS-RNN-CF  \\ \bottomrule \bottomrule
 Movies                                  & 0.594 & 0.804& 0.827 &0.909  & 0.785&  0.794  &{\bf 0.910} & 0.906  & 0.909  \\ \hline
 Music                                   & 0.533 & 0.468& 0.807 &0.802  & 0.724&  0.741  &{\bf 0.828} & 0.800  & 0.813   \\ \hline
 Travel                                  & 0.445 & 0.863& 0.853 &0.720  & 0.801&  0.824  &0.833       & 0.875  & {\bf 0.902}  \\ \hline
 Animals                                 & 0.425 &0.276 & 0.482 &0.780  & 0.603&  0.621  &{\bf0.812}  & 0.702  & 0.681  \\ \hline
 News                                    & 0.414 & 0.164& 0.466 &0.741  & 0.518&  0.555  &{\bf 0.742} & 0.543  & 0.584  \\ \hline
 Sports                                  & 0.232 & 0.000& 0.316 &0.621  & 0.523&  0.544  &{\bf 0.663} & 0.645  & 0.608  \\ \hline
 Entertainment and Cars                  & 0.307 & 0.752& {\bf 0.949}   & 0.651& 0.831&  0.856 &0.855 & 0.881  & 0.928  \\ \hline
 Games                                   & 0.321 & 0.010& 0.405 &0.748  & 0.572&  0.605  &{\bf 0.751} & 0.689  & 0.659  \\ 
  \hline\hline
 {\bf Micro-Averaged Accuracy}  & 0.450 & 0.519& 0.640 &0.793(+23\%) & 0.669(+5\%)&  0.693(+8\%)  &{\bf 0.819(+27\%)} & 0.754(+18\%)  & 0.765(+20\%)   \\ \hline 
 {\bf Macro-Averaged Accuracy}  & 0.408 & 0.482& 0.639 &0.746 (+16\%)   & 0.668(+4\%)&  0.692(+8\%)  &\bf 0.799(+25\%) & 0.755(+18\%) & 0.760(+19\%) \\
 \bottomrule
\end{tabular}
\caption{Accuracy of the topic suggestion methods compared: Popularity, CF, and Contextual-CF (C-CF) methods, vs. CTS-CRF and CTS-CNN (model-based), vs. CTS-CRF-CF and CTS-CNN-CF (hybrid models). All the results are reported for a window of size five for contextual models. All the improvements are reported based on the strongest baseline, Contextual-CF, where they are statistically significant using a one-tailed Student's t-test with p-value < 0.05.}
\label{TABLE: CRF Topic suggestion}
\vspace{-0.8cm}
\end{table*}

To evaluate our approach, we computed the topic suggestion models on off-line data for each of the methods compared. Following the established recommender system research, we used the following metrics:
\begin{itemize}
    \item \textbf{Micro-averaged Accuracy}: The accuracy is averaged across each topic suggestion individually, thus prioritizing more popular topics and potentially longer conversations.
    \item \textbf{Macro-averaged Accuracy}: The accuracy is averaged across each topic class, equally weighing both popular and ``tail'' topics.
\end{itemize}

\vspace{-0.5cm}
\subsection{Ground Truth Labels}

To create the ground-truth labels, two different scenarios have been followed for training and test data. 

For {\em training}, if a topic $t$ was suggested in turn $i$, and the talks about topic $t$ in turn $i+1$, the label of $T\_accept$ is assigned to turn $i$. If the user rejects the suggestion, or asked for something else, the label was $T\_reject$. Otherwise the label is a $follow-up$ if a user continues to engage with the same topical component, or $chat$ if the utterance is classified as non-informational or $phatic$. 

At {\em test time}, the ground truth labels were assigned as follows: if at turn $i$, a user rejects the suggested topic $T$ and subsequently, in turn $(i+n)$, requests topic $T$, then the label for turn $i$ is modified from $T\_reject$ to $T\_accept$, because it ultimately matched the user interests. Only the turns with $T\_accept$ labels were used as ground truth labels, because users accepted those suggestions at some point during the conversation. Other turns, without a true (accepted) topic, were not used for evaluation. The same ground truth labels were used for all the baseline and the proposed methods. 

\vspace{-0.4cm}

\subsection{Training CTS-CRF Model }

To train both Seq-CTS-CRF and CTS-CRF-CF models, a maximum likelihood algorithm is applied, where the parameters are optimized using Limited-memory Broyden-Fletcher-Goldfarb-Shanno (L-BFGS) method. For both methods, the context of length five turns is considered. In addition, elastic net $(L1+L2)$ regularization is used to avoid overfitting. Finally, a grid search is deployed to find the optimal values for $L1$ and $L2$, where the values of $0.03$ and $0.01$ are assigned to $L1$ and $L2$, respectively.

To implement the CF method, we trained the KNN model on the training set with a $K$ value of 33, and the cosine similarity is used as a measure of similarity. 

\vspace{-0.4cm}
\section{RESULTS AND DISCUSSION}
\label{sec:results}
We first report the main results of evaluating our proposed topic suggestion models against the popularity-based, CF, and Contextual-CF baselines. We then analyze the recommendation performance for different conversation settings and discuss some limitations of the reported experiments. 

\vspace{-0.2cm}
\subsection{Main Results}
\label{sec:main-results}

The main results on most popular classes are reported in Table~\ref{TABLE: CRF Topic suggestion}. 
The proposed CTS models outperform the strongest baseline Contextual-CF method, where CTS-CRF, CTS-CNN, CTS-RNN, CTS-CRF-CF, CTS-CNN-CF, and CTS-RNN-CF outperform the Contextual-CF method by 23\%, 4\%, 8\%, 27\%, 18\%, and 20\% respectively.

The results show that CRF significantly outperforms CNN and RNN models, which is surprising for a sequence tagging problem. RNN-based models typically outperform CRF-based methods in similar tasks like entity tagging \citep{namedentity2016}. We conjecture that CRF outperforms RNNs on this task due to two main reasons: first, the available dataset is relatively small compared to standard entity recognition datasets such as DBpedia \citep{mendes2011dbpedia} and OntoNotes 5.0 \citep{onenote2013} with more than 1200K and 1600K samples, respectively. Second, random transitions (e.g., due to dialogue breakdowns) in conversations are more frequent compared to conventional, coherent text. Users usually do not follow a standard conversation with the bot, and may randomly jump between topics. Therefore, even more data are needed to properly model the sequences. However, in contrast to deep RNNs, CRF models need significantly fewer data to be trained.

\begin{table}[ht]
\small
\begin{center}
\begin{tabular}{c|c|c|c|c}
\bottomrule
 Dropout &Num Filters&\#Hidden States & Batch Size&Accuracy\\ \bottomrule
        0.5&   128  & 100   &  64& 0.765\\   
        
        0.25&  128  & 100   &  64& 0.762(-0.0\%)\\
        0.5&   512  & 100   &  64& 0.774(+1.2\%)\\
        0.5&   128  & 300   &  64& 0.770(+0.7\%)\\
        0.5&   128  & 100   &  16& 0.764(-0.0\%)\\
        \bottomrule

\end{tabular}
\end{center}
\caption{Macro-averaged accuracy for CTS-CNN-CF with different parameter settings.}
\label{tab:pram-tuning}
\vspace{-0.8cm}
\end{table}

\begin{table*}[htb]
\begin{center}
\begin{tabular}{l|c|c|c|c|c}
\bottomrule
Method                          &   Context Size & Without Features& Topical Features & User Profile Features& All Features \\ \bottomrule\bottomrule
                       
                       CTS-RNN                  &1&    0.563  & 0.612(+8.7\%)  &  0.601(+6.7\%)   &0.665(+18.1\%)\\
                       CTS-RNN                  &3&    0.584(+3.7\%)  & 0.638(+13.3)  &  0.621(10.3\%)   &0.685(+21.6\%)\\
                       CTS-RNN                  &5&    0.613(+8.8\%)  & 0.674(+19.7\%)  &  0.656(16.5\%)   &0.693(+23.0\%)\\
                       \hline
                       CTS-RNN-CF               &5&    \bf{0.701(+24.5\%)}  & \bf{0.736(+30.7\%)}  &  \bf{0.717(+27.3\%)}   &\bf {0.765(+35.8\%)}\\
                       \bottomrule
\hline
\end{tabular}
\end{center}
\caption{Ablation study on different features of the CTS-based models, where Context Size is measured in conversation turns. All the improvements are reported based on CTS-RNN with no dialogue manager state information feature and no context. The Macro-averaged accuracy is reported, and all of the improvements are statistically significant with p-value < 0.05.}
\label{tab:ablation}
\vspace{-0.5cm}
\end{table*}

In general, the collaborative filtering approach appears to perform worse than the other models, including the Popularity-based heuristic baseline (which was manually tuned to optimize the experience of the majority of users). However, incorporating contextual information into the prediction process with CF improves accuracy by 23\%. Contextual-CF produces the best results on \textit{Entertainment and Cars}, while it is among the worst results on the other topics like \textit{Games} and \textit{Animals}. We conjecture that this is because \textit{Entertainment and Cars} is a tail topic that few users chose to engage with, and CF is designed to work well for users with rare preferences.

Similar to Contextual-CF, CTS-CNN, CTS-RNN, and CTS-CRF can effectively capture each specific conversation context, for dramatically more accurate recommendations. In contrast, they are reliable, where they provide high accuracy in all the classes. Interestingly, a hybrid of CTS (model-based) and CF model resulted in a more effective model for the topic suggestion, where CTS-CRF-CF and CTS-CNN-CF boost performance by 4\% and 9\% respectively compared to the CTS-CRF and CTS-CNN models.

\vspace{-0.3cm}
\subsection{Feature Ablation on CTS}
\label{sec:featur ablatiob}

CTS-based methods are complex models consisting of different steps built based on deep learning algorithms like CNN and RNN. We performed a comprehensive feature ablation analysis to evaluate the effect of each feature group on the overall performance of the system. Table \ref{tab:ablation} reports the results. Using all the CF, topical, and user profile features in combination, is the most effective approach for CTS-based models. Moreover, the results indicate that the impact of \textit{Topic and Behavior} is higher than \textit{User Profile} information. We conjecture that the \textit{Topic and Behavior} features contain contextual information from previous utterances. Also, as conversations progress, the values of these features are updated for each user. In contrast, \textit{User Profile} information contains static and global information about users, which remain largely unchanged during the conversation, thus having a lower impact.
\vspace{-0.2cm}
\subsubsection*{\bf Parameter Tuning. }
To evaluate how parameter tuning contributes to the final results, we performed several experiments with different parameter settings. Table \ref{tab:pram-tuning} shows macro-averaged accuracy of the CTS-CNN method with different parameters.

\vspace{-0.3cm}
\subsection{Discussion}
We now discuss the strengths and potential limitations of the proposed CTS models on different topics at different stages in the conversation. 
Finally, we provide the limitations that we encountered during our experiments.

\vspace{-0.2cm}
\subsubsection*{\bf User Topic Acceptance Rate. }
\begin{figure}
\centering
\includegraphics[width=250 pt]{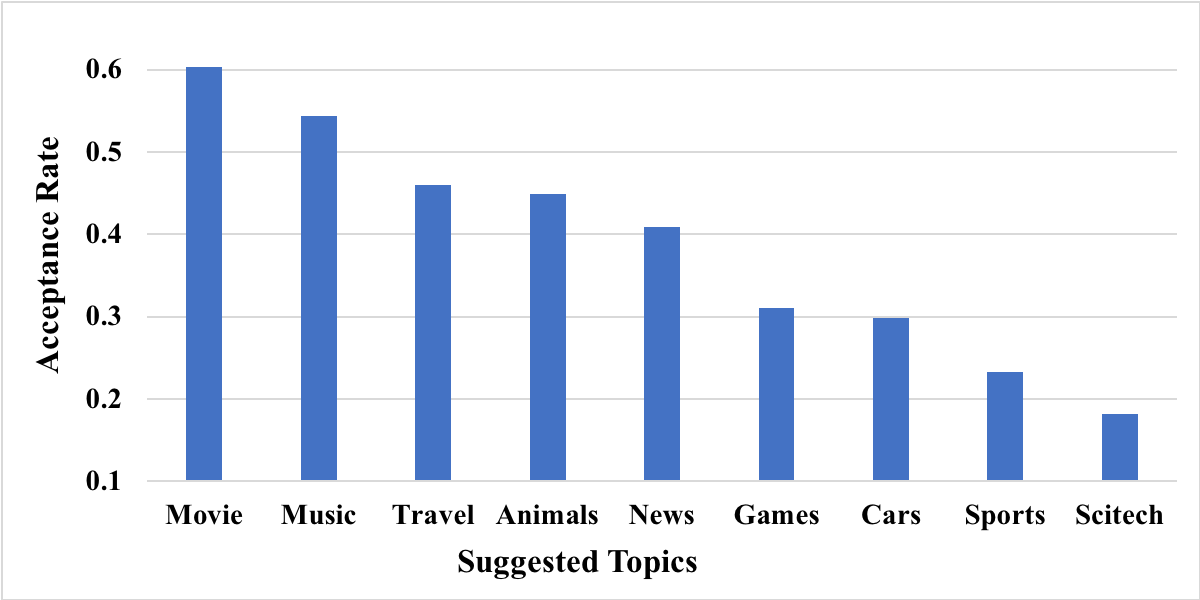}
\caption{Topic acceptance rates in Alexa dataset.}
\label{fig:dist}
\vspace{-0.4cm}
\end{figure}

Some topics are more popular and interesting for users, such as {\em Movie} and {\em Music}. The Popularity baseline described in Section \ref{sec:popularity} is designed based on these metrics. Figure \ref{fig:dist} shows the topic acceptance rate for the most popular topics in Alexa dataset. The results indicate that {\em Movie} is the most popular topic among users with over 60\% acceptance rate, and {\em Scitech} is the least favorite topic with an acceptance rate less than 20\%. 

\vspace{-0.2cm}

\subsubsection*{\bf Analyzing CF contribution to RNN- based models. }

RNN-based methods are known for finely capturing the contextual information within a sequence. The results in Table \ref{TABLE: CRF Topic suggestion} represents that using CF features contributed to the CTS-RNN by extracting relevant knowledge from the dataset that is hidden to CTS-based models. In our specific case, we can elaborate on two reasons, 1) CF features are generated using all the conversation context, while the LSTM model generally considers the history window of size $m$, and 2) CF features utilize the user-level information like the similarity between user behaviors in accepting or rejecting topics whereas RNN does not consider the user-level information.

\vspace{-0.2cm}
\subsubsection*{\bf Performance for Different Conversation Stages. }

\begin{figure}
\centering
\includegraphics[width=230 pt]{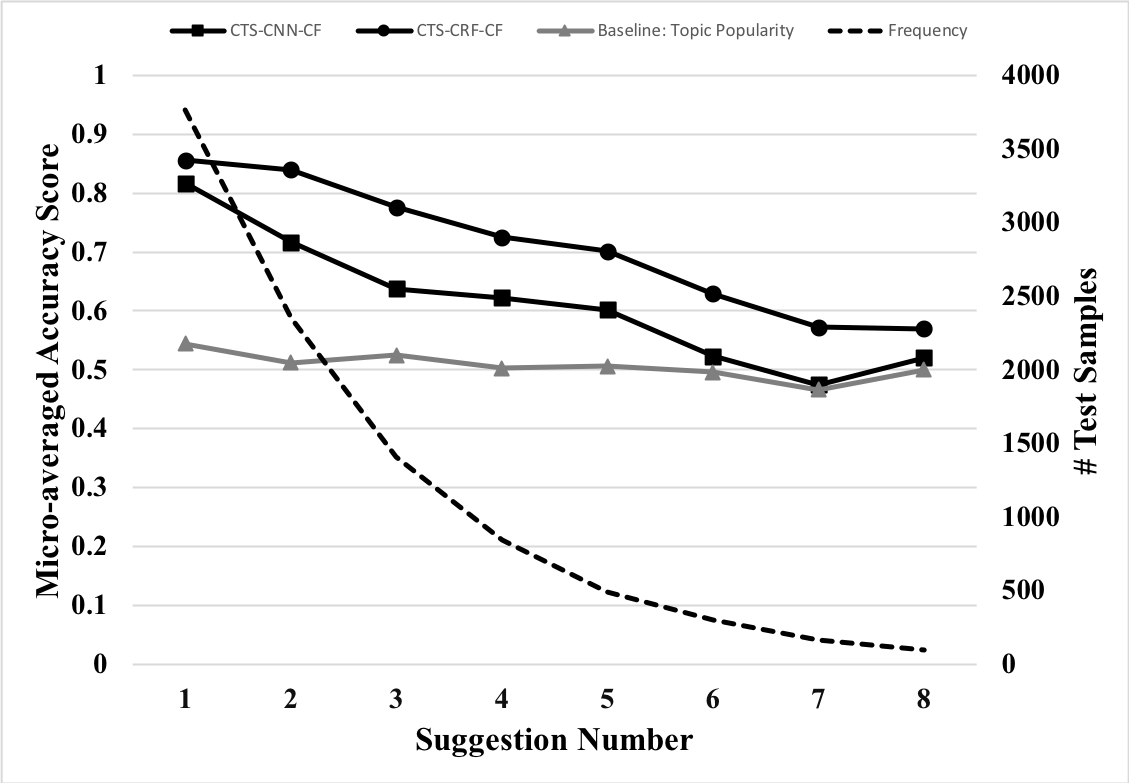}
\caption{Micro-averaged accuracy for CTS-CRF-CF, CTS-CNN-CF, and baseline Popularity model vs. the number of topic suggestions in the conversation.}
\label{fig:microresult}
\vspace{-0.6cm}
\end{figure}

As the conversation progresses, the next topic suggestion becomes increasingly challenging, as it is challenging to keep people engaged for long conversations. A proper topic suggestion model could encourage a user to engage more with the conversational agent, which has been shown to be associated with an increase in user satisfaction~\citep{choi2019offline, Venkatesh:2017}. Figure \ref{fig:microresult} reports Micro-averaged accuracy for CTS-CRF-CF, CTS-CNN-CF, CTS-RNN-CF, and the baseline Popularity model for a varying number of suggested topics per conversation. Surprisingly, the average accuracy of the suggested topic {\em drops}, as the number of suggestions in a conversation increases. 
We conjecture that this effect is due to a design decision where a direct topic suggestion was only invoked if a user was not engaged with the current topic or a domain-specific component has returned conversation's control back to the main dialogue manager. These situations indicate that the user may already be not sufficiently engaged in continuing a conversation with the conversational agent past the suggestion point. Also, the rejection of the proposed topic may not be (solely) due to the recommendation algorithm but as a result of user fatigue, or other factors. 
At the same time, fewer people continue talking to the conversational agent for the increased number of suggestions. The vast majority of people only interacting with the first one or two suggested topics. Thus, the accuracy of the first handful of suggestions is critical for user experience, as an incorrect first suggestion may cause the user to end the conversation immediately. 

\vspace{-0.2cm}
\subsubsection*{\bf Limitations. }

Our experimental evaluation used off-line analysis, and the results might differ in the on-line setting. As such, we plan to explore it in a follow-on live user study. However, we do not anticipate that the conclusions would change: we emphasize that our reported results are a {\em lower bound} on performance since we rely on conversations continuing beyond the current turn in order to give ``credit'' to our proposed suggestions that were not recommended at the appropriate time during the live competition. Another potential limitation is the form of the recommendations themselves. In this study, the system proposed a general topic like \textit{Sports} for some topics. Still, we found that proposing a specific item for the topic, e.g., ``News about the Yankees'' instead of just \textit{Sports} may be more effective, and would be a promising complementary direction to the current work, initially explored in~\cite{sahijwani2020suggestions}. 
 
\vspace{-0.2cm}
\section{CONCLUSIONS AND FUTURE WORK}

We introduced and formalized the problem of conversational topic suggestion for mixed-initiative open-domain conversational agents, specifically designed to deliver relevant and interesting information to the user. We presented and explored three approaches to this problem, a collaborative filtering based approach, a model-based sequential topic suggestion model (CTS-Seq), implemented using CRF, CNN and RNN models, and a hybrid model which combined sequence modeling approach with traditional collaborative filtering methods (CTS-CRF-CF and CTS-CNN-CF). Topic sequence models we introduced demonstrated significant improvements over previous methods, by incorporating collaborative filtering signals derived from the previous choices of similar users. 
We showed that contextual, sequence-based recommendation significantly outperforms a heavily tuned, popularity- and time-based baseline, which does not take into account the current conversation context and prior user preferences even if available.

A promising direction for future work is to further explore representation learning and deep learning techniques, to more effectively model the conversation context, and to experiment with reinforcement learning-based methods for exploring more rapidly the space of interests for new users during the conversation. These approaches would naturally fit into and extend the CTS models proposed in this paper, ultimately enabling more intelligent and proactive conversational agents.

\noindent\textbf{Acknowledgments}: We gratefully acknowledge the financial and computing support from the Amazon Alexa Prize 2018.

\balance
\vspace{-0.2cm}
\bibliographystyle{ACM-Reference-Format}
\bibliography{Reference}

\end{document}